\documentclass[journal]{IEEEtran}
\usepackage{times}
\usepackage{epsfig}
\usepackage{graphicx}
\usepackage{amsmath}
\usepackage{amssymb}
\usepackage{color}

\usepackage{subfigure}
\usepackage{multirow}
\ifCLASSINFOpdf
\else
\fi

\begin{document}
%
\title{Deep Crisp Boundaries: \\
From Boundaries to Higher-level Tasks }
%
%
%

\author{Yupei~Wang,~\IEEEmembership{Student Member,~IEEE,}
 Xin~Zhao,~\IEEEmembership{Member,~IEEE,}
 Yin~Li,~\IEEEmembership{Member,~IEEE,}

   and~Kaiqi~Huang,~\IEEEmembership{Senior Member,~IEEE}
\IEEEcompsocitemizethanks{

\IEEEcompsocthanksitem Y. Wang and X. Zhao are with the Center for Research on Intelligent System and Engineering, Institute of Automation, Chinese Academy of Sciences, Beijing 100190, China, and also with the University of Chinese Academy of Sciences, Beijing 100049, China. E-mail: wangyupei2014@ia.ac.cn, xzhao@nlpr.ia.ac.cn.\protect\
\IEEEcompsocthanksitem Y. Li is with the Department of Biostatistics and Medical Informatics and the Department of Computer Sciences, Univeristy of Wisconsin--Madison. E-mail: yin.li@wisc.edu.
\IEEEcompsocthanksitem K. Huang is with the Center for Research on Intelligent System and Engineering and National Laboratory of Pattern Recognition, Institute of Automation, Chinese Academy of Sciences, Beijing 100190, China, and also with the University of Chinese Academy of Sciences, Beijing 100049, China, and the CAS Center for Excellence in Brain Science and Intelligence Technology, 100190 (kqhuang@nlpr.ia.ac.cn).
}


}
%
%

\markboth{IEEE Transactions on Image Processing,~Vol.~xx, No.~x, October~2018}{}%

\maketitle

\begin{abstract}
Edge detection has made significant progress with the help of deep Convolutional Networks (ConvNet). These ConvNet based edge detectors have approached human level performance on standard benchmarks. We provide a systematical study of these detectors' outputs. We show that the detection results did not accurately localize edge pixels, which can be adversarial for tasks that require crisp edge inputs. As a remedy, we propose a novel refinement architecture to address the challenging problem of learning a crisp edge detector using ConvNet. Our method leverages a top-down backward refinement pathway, and progressively increases the resolution of feature maps to generate crisp edges. Our results achieve superior performance, surpassing human accuracy when using standard criteria on BSDS500, and largely outperforming state-of-the-art methods when using more strict criteria. More importantly, we demonstrate the benefit of crisp edge maps for several important applications in computer vision, including optical flow estimation, object proposal generation and semantic segmentation.
\end{abstract}


%
\IEEEpeerreviewmaketitle

\section{Introduction}
\label{sec:introduction}
\IEEEPARstart{E}{dge} detection is a well-established problem in computer vision. Finding perceptually salient edges in natural images is important for mid-level vision~\cite{martin2004learning}. Moreover, edge detection outputs, in terms of boundary maps, are often used for other vision tasks, including optical flow~\cite{revaud2015epicflow}, object proposals~\cite{APBMM2014} and object recognition~\cite{dalal2005histograms}. We have witnessed a significant progress on edge detection, ever since our community embraced a learning based approach~\cite{dollar2006supervised}. In particular, state-of-the-art methods~\cite{HED,kokkinos2015pushing}, such as Holistic Edge Detector~\cite{HED} (HED), achieved human level performance on standard datasets, e.g., BSDS500~\cite{contour}.

\begin{figure}[t]
\begin{center}
	\subfigure[]{
    \includegraphics[width=0.37\linewidth]{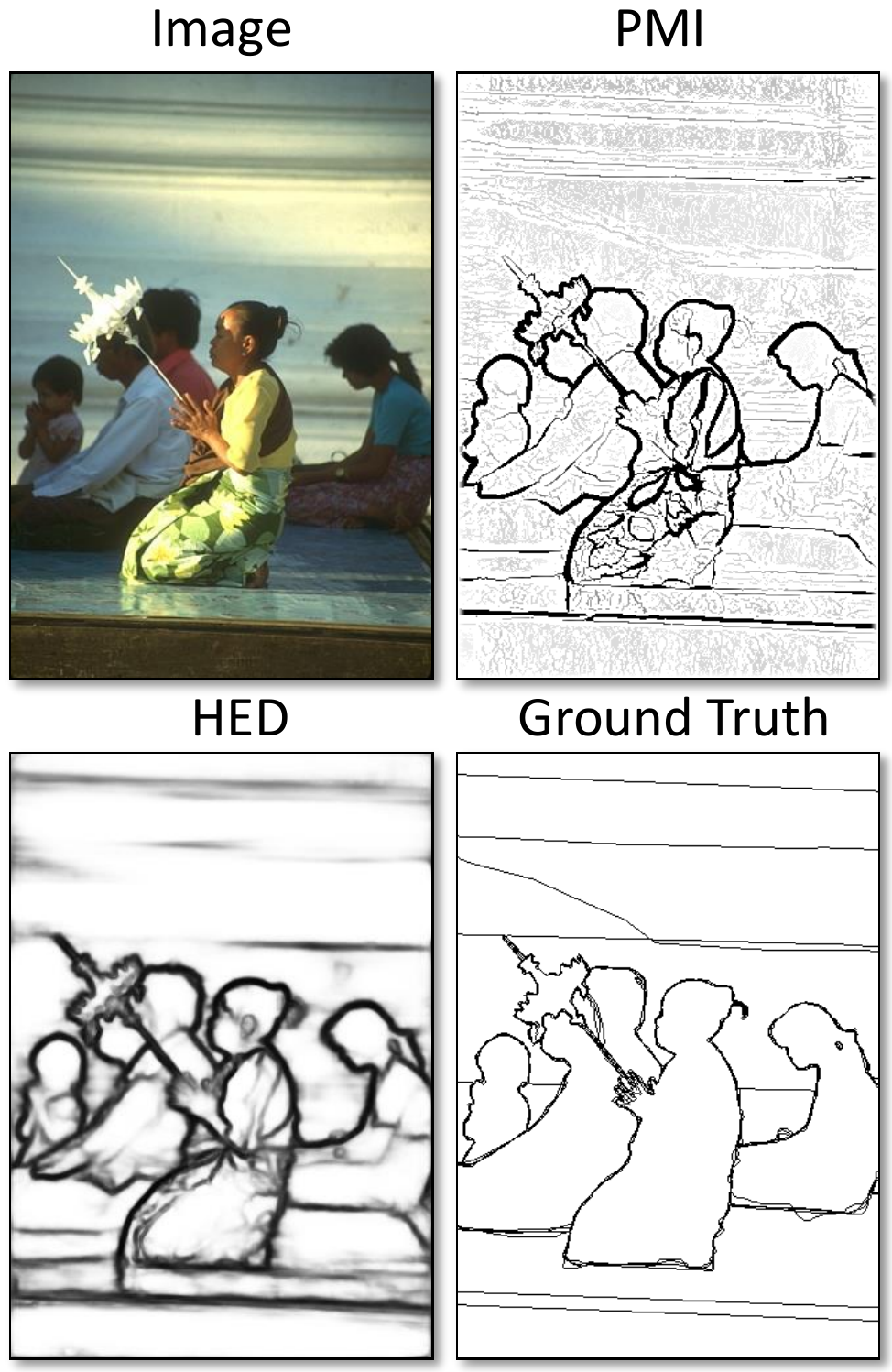}
	}
   	\subfigure[]{
    \includegraphics[width=0.52\linewidth]{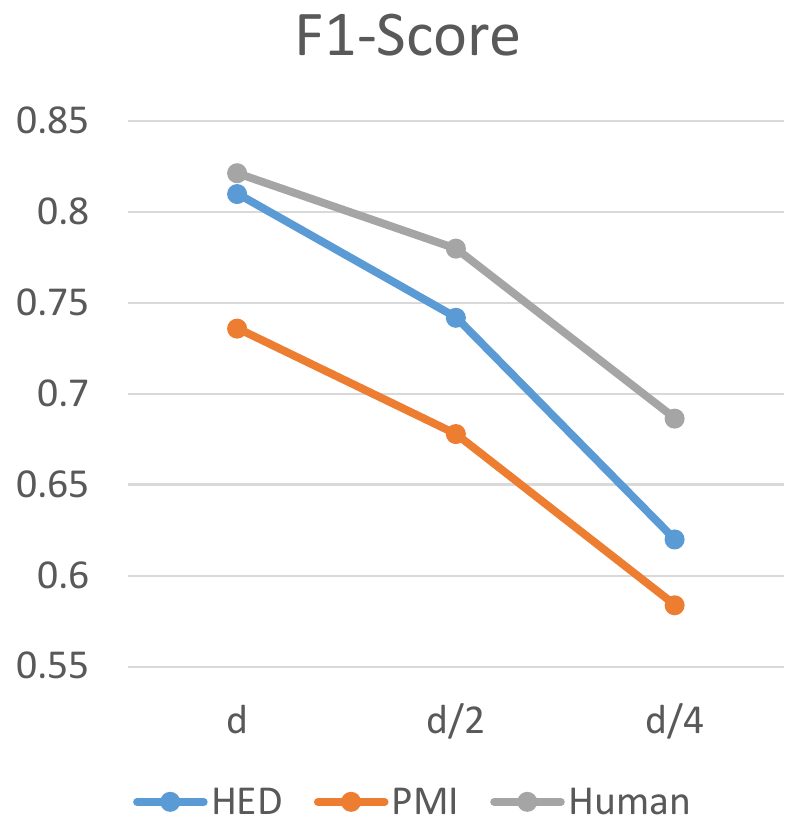}
	}
\vspace{-0.6em}    
\end{center}
   \caption{(a) Visualization of an input image, and the corresponding edge maps from PMI~\cite{crisp}, HED~\cite{HED} and human annotations. Edge map from HED is more blurry and did not precisely localize the edge pixels. (b) Performance (on the left image) drops with decreased matching distance. With a tighter criteria, the gap between PMI and HED decreases and the gap between HED and human increases. These results suggest that edge outputs from HED are not well aligned with image boundaries. We seek to improve the localization ability of ConvNet based edge detector in this paper.}
\label{fig:teaser}
\end{figure}

{\it Is edge detection a solved problem?} In Fig.~\ref{fig:teaser}(a), we show a visualization of human labeled edges, in comparison to outputs from HED (the current state-of-the-art) and PMI (designed for accurately localizing edges). While the HED result has a higher score, the quality of the edge map is less satisfactory---edges are blurry and do not stick to actual image boundaries. An accurate edge detector has to balance between ``correctness'' of an edge (distinguishing between edge and non-edge pixels) and ``crispness'' of the boundary (precisely localizing edge pixels)~\cite{crisp}. We can evaluate the ``crispness'' by decreasing the maximal permissible distance when matching ground-truth edges during benchmark. When we tighten the evaluation criteria~( the maximal permissible distance decreases from $d$ to $d/4$ ), the gap of F1 scores between HED and human increases and the gap between HED and PMI decreases~(see Fig.~\ref{fig:teaser}(b)). This result suggests the HED did not capture the precise spatial location of edge pixels.

Both qualitative and quantitative results show that edge maps from a ConvNet are highly ``correct'' yet less ``crisp''---edges are not well localized. This issue is deeply rooted in modern ConvNet architecture~\cite{krizhevsky2012imagenet}. First, spatial resolution of features is drastically reduced in more discriminative top layers due to the successive pooling layers, leading to blurred output of edges. Second, fully convolutional architecture encourages similar responses of neighboring pixels, and thus may fail to produce a thin edge map. Such a thick and blurred edge map can be adversarial for other vision tasks~\cite{crisp}. For example, recent optical flow methods~\cite{ren2008local,revaud2015epicflow} require accurate and crisp edge inputs to interpolate sparse matching results, and thus may have sub-optimal performance with blurry edges.

We address this challenging problem of learning a crisp edge detector using ConvNet, and seek to improve the localization ability of ConvNet based edge detectors. To this end, we propose a novel refinement architecture, inspired by the recent advance in dense image labeling~\cite{refine,shi2016real}. Specifically, our Crisp Edge Detection (CED) network is equipped with a top-down backward-refining pathway, which progressively increases the resolution of feature maps using efficient sub-pixel convolution~\cite{shi2016real}. This refinement pathway also adds additional non-linearity to the network, further reducing the correlation between edge responses within neighboring pixels. Our method achieves superior results on BSDS500, surpassing human performance when using standard criteria, and largely outperforming state-of-the-art methods when using more strict evaluation criteria. Moreover, we explore variations of CED by using different backbone networks~\cite{simonyan2014very,He2015Deep} and adding additional convolutional layers with large kernel size~\cite{Hou2016Deeply}. Our improved version of CED established new state-of-the-art results on BSDS500~\cite{contour} and PASCAL-Context~\cite{mottaghi2014role} datasets for the task of edge detection.

More importantly, we demonstrate that our crisp edge outputs can positively improve the performance of several mid-level or high-level vision tasks. Specifically, we focus on three important tasks, including optical flow estimation, object proposal generation and semantic segmentation. We show that well-localized boundaries can help for (1) more accurate estimation of boundary-preserving optical flow; (2) more precise generation of object proposals that stick to object boundaries; and (3) better localized object masks.

Our contributions are thus summarized into three parts.
\begin{itemize}
  \item We provide a systematical study of edge maps from ConvNet. And we show that previous models are good at classifying edge pixels yet have poor localization ability.
  \item We propose a novel deep architecture tailored for learning crisp edges. Our method combines the refinement scheme~\cite{refine} with sub-pixel convolution~\cite{dollar2006supervised}. Our results on BSDS500 and PASCAL-Context outperform state-of-the-art methods on a large range of matching distances.
  \item We show that crisp edge maps can improve mid-level and high-level vision tasks, including optical flow estimation, object proposal generation, and semantic segmentation.
\end{itemize}

An early version of this work appeared in~\cite{yupei2017deepcrisp}. And we have made substantial extensions to our previous work. Specifically, we further improve the model by exploring deeper networks. We also include more experiments and analysis on edge detection, and add a new experiment on semantic segmentation. Finally, our paper is organized as follows. Section~\ref{sec:Survey} reviews related work on edge detection. Section~\ref{sec:Thick} presents our study of edge maps from ConvNet. Section~\ref{sec:Method} details our method. Section~\ref{sec:edge detection} demonstrates experimental results for boundary detection. Section~\ref{sec:benefits} shows the benefits of crisp boundaries.

\section{Related Work}
\label{sec:Survey}

\subsection{Boundary Detection}
There is a vast literature on the classical problem of edge detection. A complete survey is out of scope for this paper. We only review a subset of relevant works in this paper.

Early edge detectors were manually designed to find discontinuities in intensity and color~\cite{fram1975quantitative,canny1986computational,freeman1991design}. Martin et al.\ \cite{martin2004learning} found that adding texture gradients significantly improves the performance. Several recent works explored learning based approaches for edge detection. Doll\'{a}r et al.\ \cite{dollar2006supervised} proposed a data-driven, supervised edge detector, where detection is posed as a dense binary labeling problem with features collected in local patches. Many modern edge detectors have followed this paradigm by using more sophisticated learning methods. For example, Ren and Bo~\cite{xiaofeng2012discriminatively} combined features learned from sparse coding and Support Vector Machine (SVM) for edge detection. Lim et al.\ \cite{lim2013sketch} proposed to cluster human generated contours into so called Sketch Tokens, followed by the learning of a random forest that maps a local patch to these tokens. These tokens are finally used to re-assemble local edges. This idea was further extended by Doll\'{a}r and Zitnick~\cite{dollar2015fast}. They proposed structured random forest that simultaneously learns the clustering and the mapping. The random forest thus directly outputs a local edge patch. However, none of these previous methods considered deep models.

The recent success of deep ConvNet has greatly advanced the performance of edge detection. For example, Bertasius et al.\ ~\cite{bertasius2015deepedge} presented a two-branch network to classify and regress edge maps from candidate image patches. Shen et al.\ \cite{shen2015deepcontour} proposed to cluster contour patches with different shape and further assemble them into contours. Going beyond low-level cues, Bertasius et al.\ \cite{bertasius2015high} exploited object-related features for boundary detection. More recently, Xie and Tu~\cite{HED} proposed to combine Fully Convolutional Networks (FCN)~\cite{long2015fully} with deep supervision~\cite{lee2015deeply}. Their method made use of features from different scales using skip-layer connections and achieved a superior performance within $2\%$ gap to human. Kokkinos et al.\ \cite{kokkinos2015pushing} further extended HED by adding multi-instance learning, more training samples and a global grouping step. Yang et al.\ \cite{ContourDetection2016} presented an encoder-decoder architecture for object contour detection. Other efforts included weakly supervised~\cite{khoreva2016weakly} and unsupervised~\cite{LiCVPR16edges} learning of edges. 

Our method adds backward refinement pathway to HED~\cite{HED}. Our network thus can be viewed as encoder-decoder as in~\cite{ContourDetection2016}, yet with specially designed skip-connections between the encoder and the decoder. Moreover, previous methods focused on the ``correctness'' of edges by selecting an optimistic matching distance ($4.3$ pixels in a resolution of $321\times 481$) and overlooked the ``crispness'' of edges. Their performance thus drops dramatically when the evaluation criteria is tightened. In contrast, our method seeks to accurately localize edge pixels and addresses the ``crispness' of edges.

Our method is motivated by Isola et al.\ \cite{crisp}. They proposed an affinity measurement based on point-wise mutual information between distributions of hand-crafted local features. Edges are then detected using this affinity by spectral clustering. We share the same goal of designing a crisp edge detector yet our method and setting are completely different. More precisely, we pursue a learning based approach using ConvNet for crisp edges. Finally, our method is inspired by Pinheiro et al.\ \cite{refine}, where a refinement architecture is proposed for segmenting objects. Our method adopts the top-down pathway of~\cite{refine} to label the sparse binary signals of edges. Yet we replace the bilinear interpolation (deconvolution) with sub-pixel convolution~\cite{shi2016real}, which is critical for generating better-localized, sharp edge outputs. More importantly, we explore the benefit of crisp edges for mid-level and high-level tasks.

\subsection{Boundaries for Vision Tasks}

Boundary detection is an important step in many mid-level and higher-level vision tasks. For example, EpicFlow~\cite{revaud2015epicflow}, a state-of-the-art optical flow method, made use of boundaries for edge-preserving interpolation from sparse matches for accurate dense matching. Moreover, several object proposal generation methods also need accurate edge prediction to create proposes that sticks to object contours (e.g., MCG~\cite{APBMM2014}). And finally, boundary cues provide fine details of objects and are thus important for semantic segmentation. For example, boundary maps can be used as a post-processing step to refine the coarse outputs from a network~\cite{Bertasius2015Semantic}. And this post-processing can be learned from end-to-end~\cite{chen2016semantic}.

In this paper, we specifically explore the benefit of crisp edges for these mid-level and high-level vision tasks. This is closely related to the work from Maninis et al.\ \cite{maninis2016convolutional}. They proposed to explicit model the orientation of edges in the deep networks. And they demonstrated that this additional channel of edge orientation can help higher level vision tasks. In the same spirit, we show that using our crisp edges can help to improve the performance of optical flow estimation, object proposal generation and semantic segmentation.

\begin{figure}[t]
\begin{center}
	\subfigure[]{
    \includegraphics[width=0.95\linewidth]{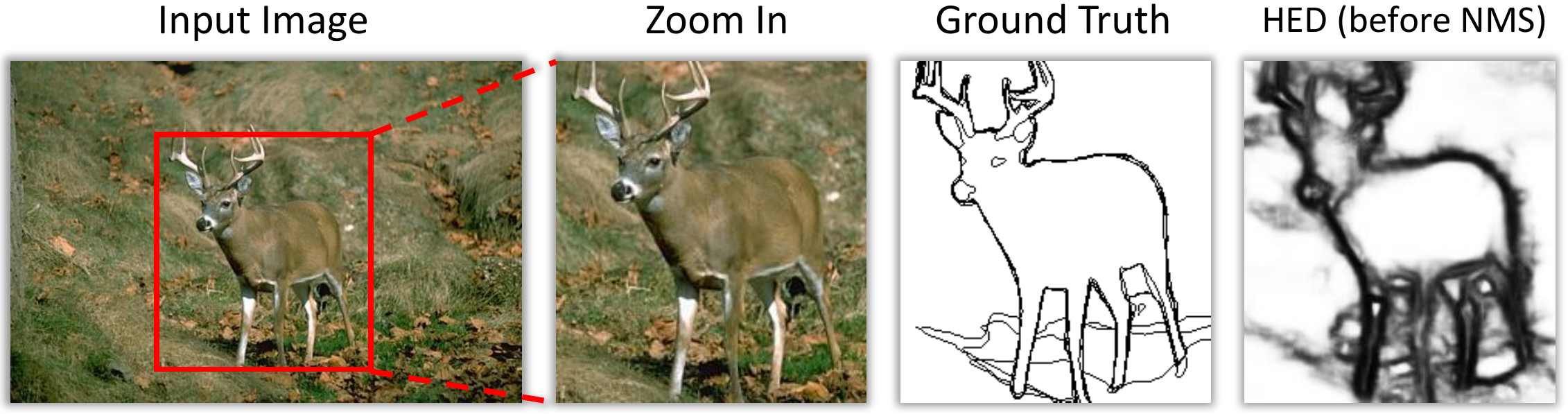}
	}
   	\subfigure[]{
    \includegraphics[width=0.9\linewidth]{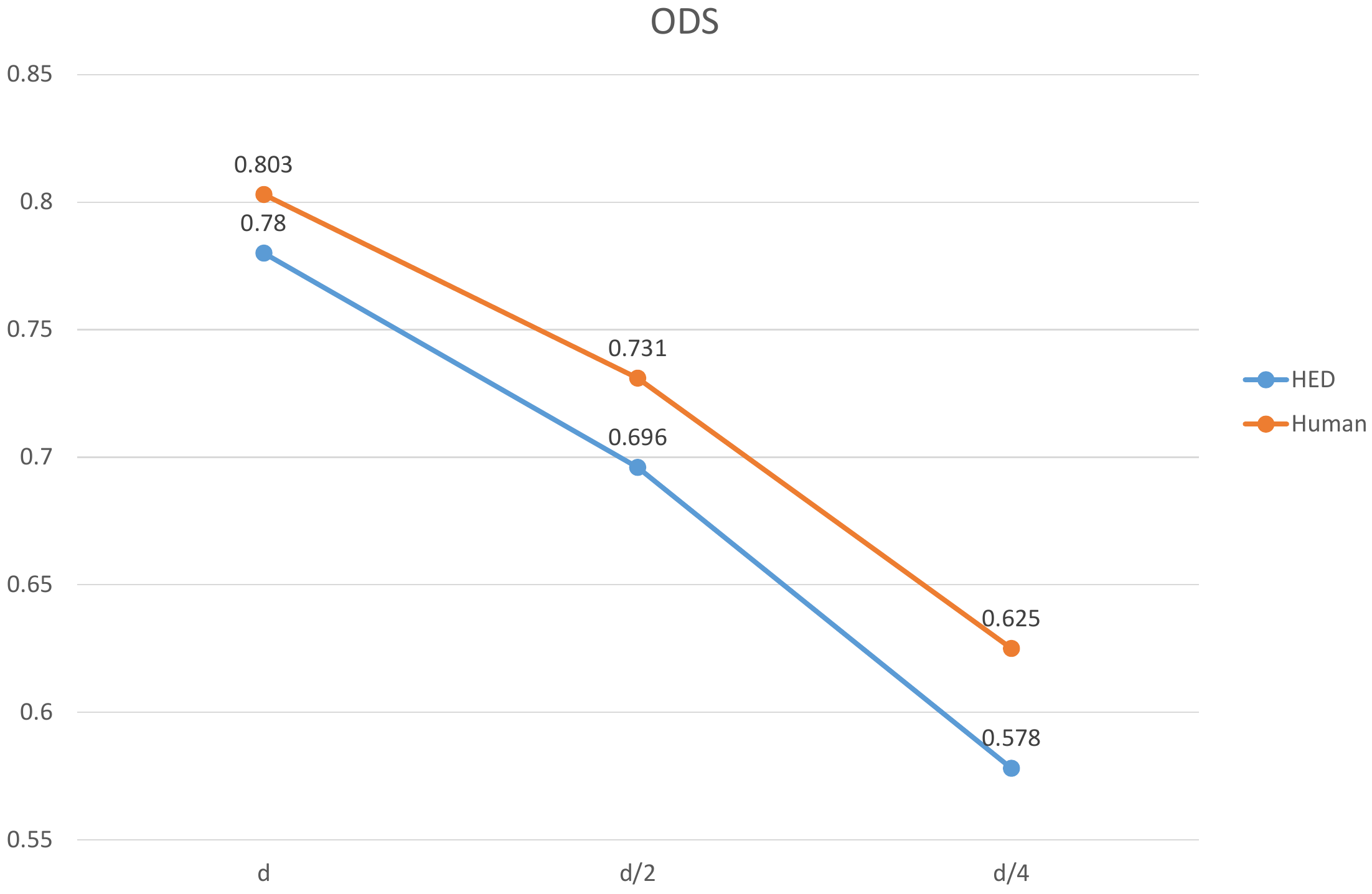}
	}
\vspace{-0.6em}
\end{center}
   \caption{(a) Thick and noisy edge map generated with HED~\cite{HED} before non-maximal suppression(NMS); (b) Optimal Dataset Score (ODS) for both HED and human drop with decreased matching distance on the BSDS500 test set. However, the performance gap between HED and human increases from $2.3\%$ to $4.7\%$ as the distance decreases from $d$ to $d/4$. }
\label{fig:HED}
\end{figure}

\section{Thick Boundaries from ConvNet}
\label{sec:Thick}

We start by looking into the output edge maps of HED~\cite{HED}, a recent successful edge detector using ConvNet. HED predicts edge confidence at different layers of the network, leading to a set of edge maps. These maps are down-sampled due to successive pooling operations in the network. Then, they are further up-sampled to fit the input resolution by bilinear interpolation and averaged to produce the final edge map. We show an example of the edge map in Fig.~\ref{fig:HED}(a). The detector achieved an ODS of $0.78$ on BSDS500. However, the visual quality of the edge map is unsatisfactory. The edge map looks blurry and visually defective.

Why would such a blurry edge map reach a high score in benchmark? The standard evaluation~\cite{contour} iterates over all confidence thresholds and uses bipartite graph matching to match between a binarized edge map to ground-truth edges. The matching is controlled by a maximal permissible distance $d$. A misaligned edge pixel is still considered correct as long as its distance to the nearest ground-truth edge pixel is smaller than $d$ pixels. With an optimistic $d$, we can achieve a high score even if edges are slightly mis-aligned.

In fact, edge detection has to balance between ``correctness'' of an edge (distinguishing between edge and non-edge pixels) and ``crispness'' of the boundary (precisely localizing edge pixels)~\cite{crisp}. A well-aligned edge map (crisp edges) can be critical for other vision tasks, such as optical flow or image segmentation. This ``Crispness'' can be measured by decreasing $d$ in the benchmark. Human performance gradually decreases with smaller $d$, as we show in Fig.~\ref{fig:HED}(b). However, HED outputs show a more drastic drop, indicating that HED edges are not well aligned to actual image boundaries. Our visual inspection of the edge map also reveals this trend, e.g., blurry edges. 

\begin{figure*}
\begin{center}
\includegraphics[width=1.0\linewidth]{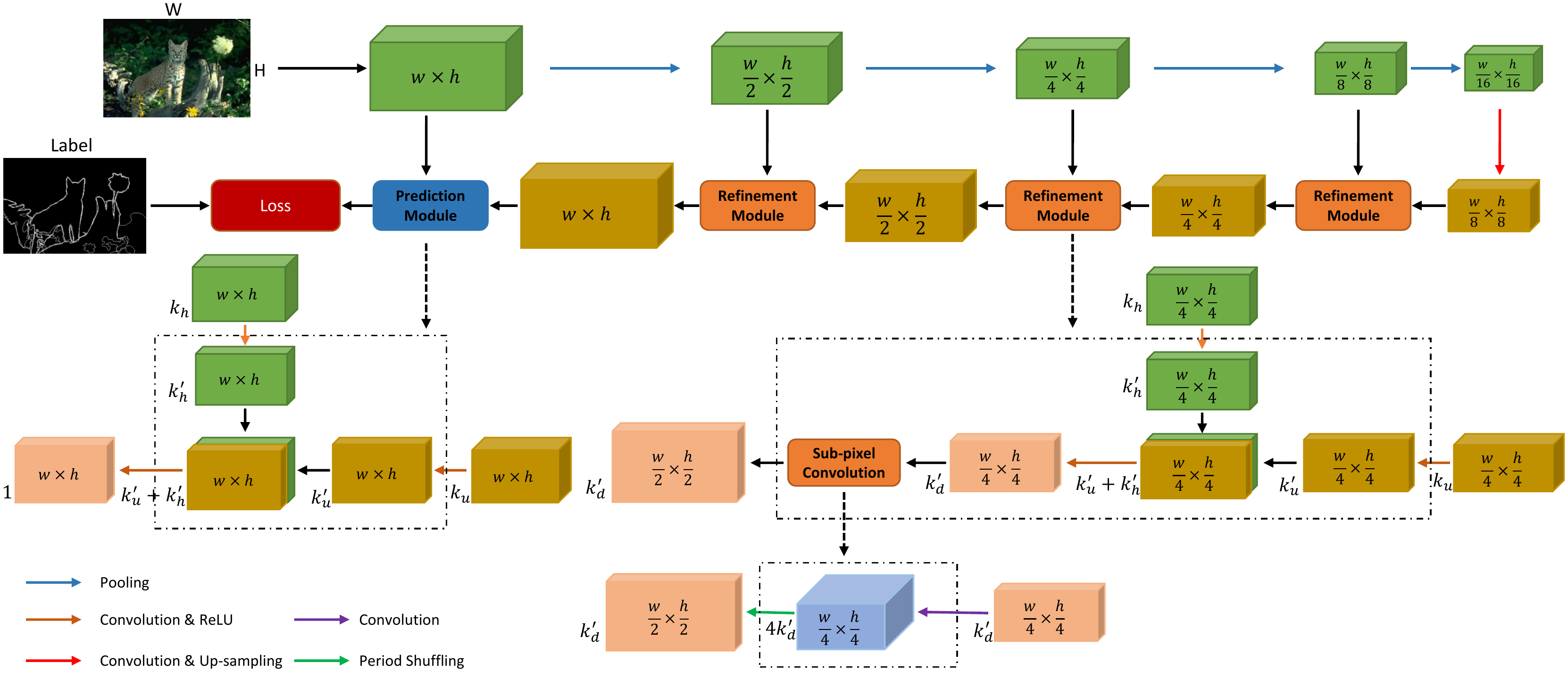}
\end{center}
   \caption{Architecture of the proposed Crisp Edge Detector (CED) network. We add a backward-refining pathway to a backbone HED network. This pathway progressively increases the resolution of feature maps. And our refinement module fuses a top-down feature map with feature maps on the forward pass, and up-samples the map using sub-pixel convolution. Our model is specially designed for generating edge maps that are well-aligned to image boundaries.}
   \label{fig:network}
\end{figure*}

\section{Make Convolutional Boundaries Crisp}
\label{sec:Method}
How can we make a crisp edge map from ConvNet? We start by analyzing the architecture of HED. Like modern ConvNets, spatial resolution of more discriminative top layers is significantly reduced due to the successive pooling operations. HED further attached a linear classifier on layers with different resolution, and uses bilinear interpolation~(realized as deconvolution) to up-sample their outputs to the original resolution. This design has two major issues. First, linear classifiers within a fully convolution architecture will produce similar responses at neighboring pixels. It is thus difficult to distinguish an edge pixel from its neighbors. More importantly, up-sampling using bilinear interpolation is not sufficient to recover spatial details, and thus further blurs the edge map.

Architecture modifications are therefore required for generating a crisp edge map. In this section, we address the challenging problem of designing a Crisp Edge Detector (CED) by proposing a novel architecture. Our method supplements HED network with a backward-refining pathway, which progressively up-samples features using efficient sub-pixel convolution~\cite{shi2016real}. Our proposed CED is able to generate an edge map that is well-aligned with image boundaries. We now present details of the proposed CED network, explain the insight of our design, and describe our implementation.

\subsection{Architecture Overview}

Fig.~\ref{fig:network} shows an overview of CED with two major components: the forward-propagating pathway and backward-refining pathway. The forward-propagating pathway is similar to HED. It generates a high-dimensional low-resolution feature map with rich semantic information. The backward-refining pathway fuses the feature map with intermediate features along the forward-propagating pathway. This refinement is done multiple times by a refinement module. Each time we increase the feature resolution by a small factor ($2$x) using sub-pixel convolution. And eventually we reach the input resolution. Details of our network are elaborated in following subsections.

\begin{figure}[h]
\begin{center}
\includegraphics[width=0.8\linewidth]{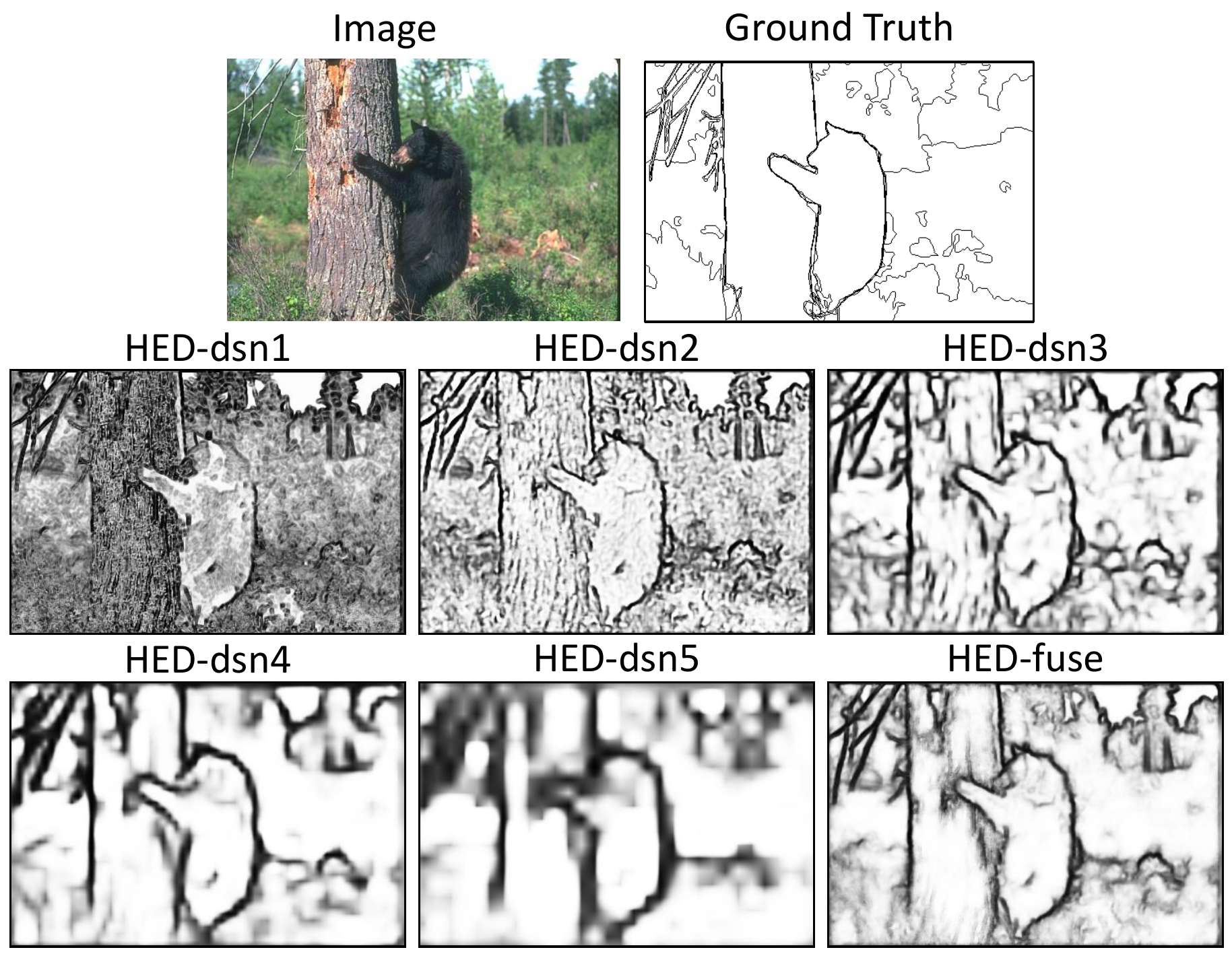}
\vspace{-0.5em}
\end{center}
   \caption{Predictions from all side outputs and the final output ~(HED-fuse) of HED network~\cite{HED}. The lower layers~(HED-dsn1/2/3) capture finer spatial details, while lack sufficient semantic cues. However, the deeper layers~(HED-dsn4/5) encode richer semantic cues, but spatial details are missing.}
   \label{fig:HED_col}
\end{figure}

\subsection{Refinement Module}

The skip-layer connection provides HED the important ability to use features at different layers for finding edges~\cite{HED}. Fig.~\ref{fig:HED_col} shows example side predictions of its five side-outputs and final edge map~(HED-fuse). The lower layers~(HED-dsn1, HED-dsn2, HED-dsn3) capture rich local spatial details, while lack sufficient semantic cues. However, the deeper layers~(HED-dsn4, HED-dsn5) encode richer global semantic cues, but spatial details are missing. HED simply fuses these independent side predictions by a weighted fusion. We argue that this is not a good design as it does not explore the hierarchical feature representations of ConvNet. To get a better fusion of the multi-layer features, we introduce the backward-refining pathway with refinement modules, similar to~\cite{refine}. Note that our task of detecting sparse edges is significantly different from segmenting objects in~\cite{refine}. Thus, directly applying the same module in~\cite{refine} leads to sub-optimal performance.

The refinement module is repeated several times to progressively increase the resolution of feature maps. The key idea is to aggregate evidences of edges across the path using intermediate feature maps. Detailed structure of the module is shown in the bottom part of Fig.~\ref{fig:network}. Each module fuses a top-down feature map from the backward pathway with the feature map from current layer in the forward pathway, and further up-samples the map by a small factor ($2$x), which is then passed down the pathway. There are two core components in this module, namely \textbf{fusion} and \textbf{up-sampling}.

\noindent \textbf{Fusion}. A straightforward strategy is to directly concatenate two feature maps. However, this is problematic since each map has a large number of channels. Directly concatenating the features will dramatically increase the number of parameters in the model. To this end, we reduce the dimension of both feature maps through additional convolutions, and concatenate the two low-dimensional feature maps with equal channels.

We denote the number of channels of the input forward pathway feature map as $k_h$. After the convolutional and ReLU operations, the channels are reduced to $k_h^{'}$, which is much less than $k_h$. The same operations are conducted to the feature map from the previous refinement module to produce $k_u^{'}$ from $k_u$. We concatenate the above feature maps into a new feature map with $k_u^{'} + k_h^{'}$ channels and reduce it to a feature map with $k_d^{'}$ channels by a $3\times3$ convolutional layer as well. Thus, the overall computational cost is reduced.

\noindent \textbf{Up-sampling}. After fusion, our refinement module will also need to expand the resolution of feature maps. We up-sample the fused feature map with a sub-pixel convolution~\cite{shi2016real}. The sub-pixel convolution, different from the bilinear interpolation for up-sampling~\cite{zeiler2011adaptive, dumoulin2016guide, vedaldi2015matconvnet}, is a standard convolution followed by additional rearrangement of feature values, termed phase shift. It has shown to help removing the block artifact in image super-resolution task and maintaining a low computational cost. We found that using sub-pixel convolution is important for accurate localization of edges.

Supposed we have $i$ input channels and $o$ desired output channels, the kernel size of a convolutional layer is denoted as $(o,i,r,c)$, where $r$ and $c$ stand for the kernel width and kernel height respectively. Considering output feature map with $k$ times larger resolution than the input one, the traditional deconvolutional layer would employ the kernel size to be $(o, i, k\times{r},k\times{c})$. Instead of directly output enlarged feature map through a single deconvolutional layer, the sub-pixel convolution consists of one convolutional layer and one following phase shift layer. The kernel size of the convolutional layer is $(o \times k^{2}, i, r, c)$, thus generating feature map with $o \times k^{2}$ feature channels with identical resolution. We then apply the phase shift to assemble the output feature map to the feature map with $o$ feature channels but $k$ times larger resolution.

\noindent \textbf{Sub-pixel Convolution vs.\ Deconvolution}. These two operations are equal only if (1) parameters in deconvolution are trainable; and (2) more parameters are added in deconvolution layers~\cite{shi2016real}. In HED, deconvolution is fixed to bilinear upsampling where its parameters are not updated. Even if we allow the learning of deconvolution, sub-pixel convolution still requires less number of parameters. Sub-pixel convolution thus provides an efficient way of learning to upsample. We found it helpful for improving the performance in our experiments. Similar results are also observed in~\cite{wang2017understanding}.

\noindent \textbf{Relationship to~\cite{HED} and~\cite{refine}}. CED subsumes HED~\cite{HED} as a special case, where $3$x$3$ convolutions and ReLUs are replaced by linear classifiers and progressive up-sampling is used. Our method is different from~\cite{refine} as we replace bilinear interpolation with sub-pixel convolution. This enables a more expressive model with a small number of extra parameters. Our task of edge detection is also different from object segmentation in~\cite{refine}.

\subsection{Further Improvements to CED}
We have made further improvements over CED model. Concretely, we looked at (1) using a better backbone network (ResNet); and (2) using larger kernels for each side output.

\noindent \textbf{Better Backbone Network}. We replace the VGG backbone used in the original HED with more recent ResNet~\cite{He2015Deep}. In comparison to VGG, ResNet has shown to provide better performance on a number of vision tasks~\cite{He2015Deep,maninis2016convolutional}. Since spatial details are critical for crisp boundary detection, we remove the first pooling operation in the network and thus the output is downsampled by a factor of 16 (similar to HED~\cite{HED}).

\noindent \textbf{Large Kernels for Side Outputs}. Furthermore, as in~\cite{Hou2016Deeply}, we improve CED by adding additional two convolutional layers at each of the five side-outputs in backbone network. This is inspired by~\cite{peng2017large} where they showed that large kernel size helps to enable dense connection between feature maps and final classifiers, and thus is conducive to cope with different transformations. We set the kernel size of the newly added layers for the five side-outputs as 3x3, 3x3, 5x5, 5x5 and 7x7.

\subsection{Insights into the CED Network}
Our goal is to increase the ``crispness'' of the output edge maps. This is realized by the design of our CED network. Specifically, we develop a novel refinement architecture, and further add sub-pixel convolution to replace deconvolution. Our architecture benefits from progressively refined feature maps that add fine details of boundaries. Moreover, we add additional non-linearity to all side outputs, which further reduces the correlation between edge responses within neighboring pixels. Finally, we use the sub-pixel convolution to up-sampling the feature map. This operation is able to remove the blurriness introduced by a traditional deconvolution. In this way, our CED network can be trained from end-to-end, and produce an edge map that is both ``correct'' and ``crisp''.

\begin{figure}[t]
\begin{center}
\includegraphics[width=1.0\linewidth]{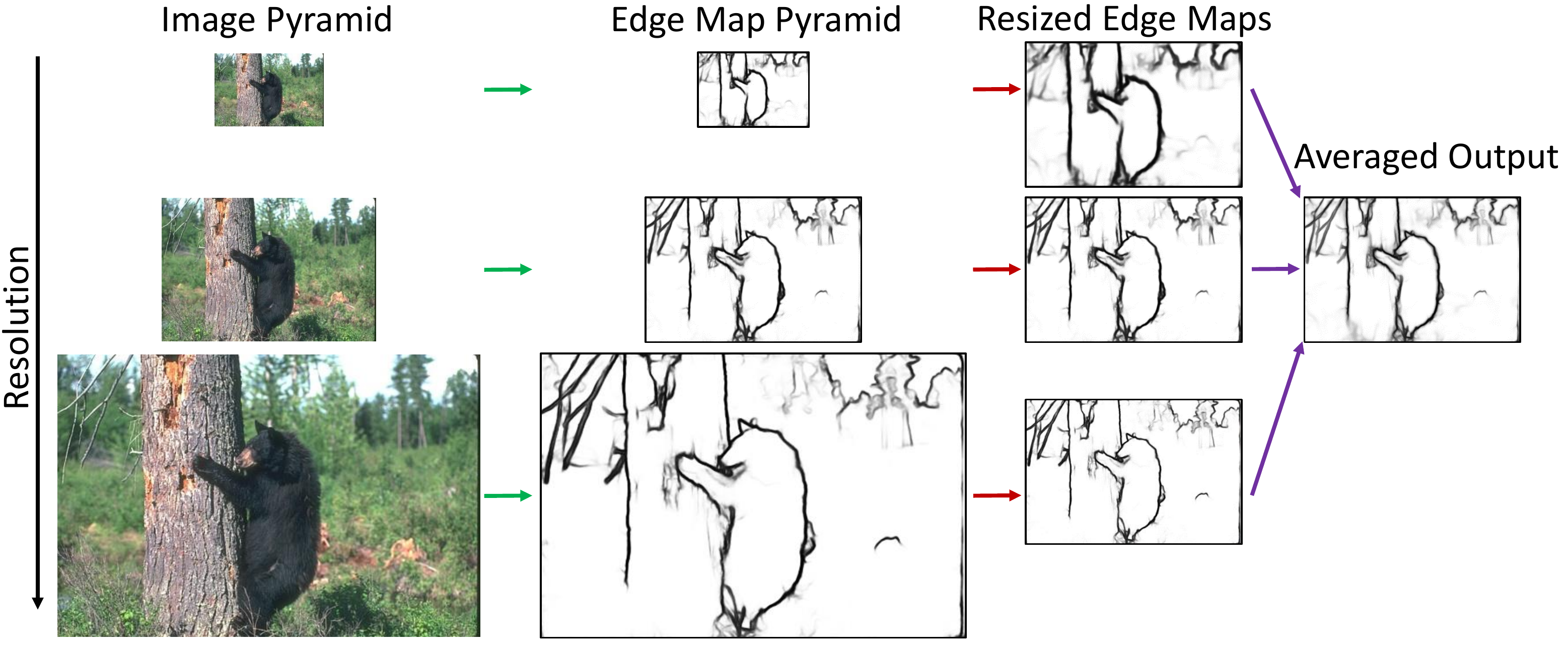}
\vspace{-0.5em}
\end{center}
  \caption{Multi-scale fusion strategy (only used for evaluation). We average outputs from multiple scales to further improve the performance.}
\label{fig:multi}
\end{figure}

\subsection{Implementation Details}
Our implementation builds on the publicly available code of HED~\cite{HED}, using Caffe as the backend~\cite{caffe}. For training, we initialize the forward-propagating pathway with the pre-trained HED model. All other layers are initialized with Gaussian random distribution with fixed mean ($0.0$) and variance($0.01$). We use back-propagation and Stochastic Gradient Descent (SGD) for training the network. Our hyper-parameters for training, including the initial learning rate, weight decay and momentum, are set to $1e$-$5$, $2e$-$4$ and $0.99$, respectively.

More concretely, for backward-refining pathway, the number of convolutional kernels is set to $256$ for the top layer. This number is decreased by half along the path. For example, the first, second, and third top-down refinement module will have $128$, $64$ and $32$ feature channels, respectively. Since the resolution of feature maps decreases by a factor of $2$ after every pooling operation, the sub-pixel convolution up-samples the input feature map by $2$x in each refinement module.

During testing, we make use of a multi-scale fusion strategy to further boost the performance. Specifically, an input image is resized to three different resolutions ($1/2$x, $1$x, $2$x), which are fed into the same network independently. We then resize the three output edge maps to the original resolution, and average them to generate the final edge map. Fig. \ref{fig:multi} shows the pipeline of the multi-scale fusion strategy for the evaluation.

\section{Crisp Edge Detection}
\label{sec:edge detection}
Our first set of experiments focuses on the task of edge detection. We conduct extensive experiments to benchmark our proposed CED. We introduce our datasets and evaluation criteria, present an ablation study of CED, and compare CED to state-of-the-art methods on two public benchmarks.

\subsection{Datasets and Benchmarks}
\label{subsec:edge detection_datasets}
We evaluate our method on the widely-used Berkeley Segmentation Dataset and Benchmark (BSDS500) dataset~\cite{martin2004learning,contour}. It consists of 200 images for training, 100 for validation, and 200 for testing. Each image is annotated by multiple annotators. We use the train and validation set for training (similar to~\cite{HED}) and report results on the test set. The performance is measured by the precision/recall curve that captures the trade-off between accuracy and noise \cite{martin2004learning}. In addition, three standard metrics are reported: fixed contour threshold (ODS), per-image best threshold (OIS) and average precision (AP).

Following~\cite{maninis2016convolutional,HED}, we also evaluate our method on the PASCAL-Context dataset~\cite{mottaghi2014role}. This dataset provides the pixel level labeling for 10103 images including both objects and stuff. Similar to~\cite{HED}, we consider the top 60 categories. Ground-truth edge map is extracted from the semantic labels. A pixel is assigned as boundary if one of its neighbors has a different label. We train our models on the training set (4998 images) and report results on the validation set (5105 images).

\begin{table}[t]
\begin{center}
\begin{tabular}{l|c|c|c}
Method&ODS & OIS&AP\\
\hline
HED & .780 & .797& .829 \\
Res16x-HED & .785 & .804& .857 \\
Res16x-HED-large & .793 & .811 & .864 \\ \hline
CED-sub-multi & .793 & .811 & .838 \\
CED-multi & .794 & .811 & .847 \\
CED-sub & .800 & .819 & .859 \\
CED & \textbf{.803} & \textbf{.820} & \textbf{.871} \\ \hline
Res16x-CED & \textbf{.810} & \textbf{.829} & \textbf{.879} \\
\end{tabular}
\end{center}
\caption{Results on BSDS500 with different network architectures. CED-w/o-Multi refers to CED without multi-scale testing, similar for CED-w/o-Subpixel-w/o-Multi. Our final model improves the baseline by $2.3\%$ when using VGG16. }
\label{tab:comp}
\end{table}

\subsection{Ablation Study}
To begin with, we conduct an ablation study for different network architectures of CED using BSDS500 dataset. Specifically, we compare the following variants of the model
\begin{itemize}
\item \textbf{HED}: This is the original HED~\cite{HED} network using VGG16~\cite{simonyan2014very}. It is used as our baseline model.
\item \textbf{Res16x-HED}: This is the HED with ResNet50~\cite{He2015Deep} as backbone. It is an enhanced version of original HED~\cite{HED}.
\item \textbf{Res16x-HED-Large}: This is the Res16x-HED with large kernels. It helps to calibrate the performance with our improvements to CED.
\item \textbf{CED-sub-multi}: This is the CED network using VGG16 without subpixel convolution and multi-scale testing. Thus, this model only adds refinement pathway to HED.
\item \textbf{CED-multi}: This is the CED network using VGG16 without multi-scale testing. This is the base of our full model with both refinement pathway and subpixel-convolution.
\item \textbf{CED-sub}: This the CED network using VGG16 without subpixel convolution. This model further adds multi-scale testing to CED-sub-multi.
\item \textbf{CED}: This is our full CED model with multi-scale testing yet with VGG16 as the backbone.
\item \textbf{Res16x-CED}: This is our improved version of CED network. This model makes use of ResNet50 as backbone, and adds large kernel convolutions for side outputs in backbone network.
\end{itemize}

\noindent \textbf{Training Protocol}. For training, we adopt a modified version of consensus sampling strategy~\cite{HED} to prevent the problematic convergence behavior. A pixel is assigned positive label if it is labeled as edge by at least three annotators. Pixels have not been labeled by any annotators are treated as negative. The rest of the pixels are ignored during training (by blocking their gradients). We also augment the data by rotating, cropping and multi-scale resizing. For fair comparison, all CED models are initialized from the same HED base model.

\noindent \textbf{Results}. Our results are summarized in Table~\ref{tab:comp}. First, our refinement pathway improves the baseline HED by $1.3\%$ in ODS score (CED-sub-multi vs.\ HED). Second, further adding sub-pixel convolution leads to a minor boost of $0.1\%$ under standard evaluation criteria (CED-sub-multi vs. CED-multi). Yet we have observed that sub-pixel convolution is critical when using a tight criteria. Third, the multi-scale testing adds another $0.7$-$0.9\%$ (CED-sub-multi vs. CED-sub or CED-multi vs. CED). Moreover, Res16x-HED improves HED by $0.5\%$, which demonstrates the effectiveness of the better backbone network. Adding additional convolutional filters with large kernel in side-outputs, Res16x-HED-Large further boosts the performance of Res16x-HED by $0.8\%$. Res16x-CED significantly outperforms Res16x-HED-Large by $1.7\%$--a strong performance gain from the architecture of CED. Finally, we note that our improved model (Res16x-CED) is a new record for single model on BSDS500 without external training data.

\begin{figure}[t]
\begin{center}
\includegraphics[width=0.9\linewidth]{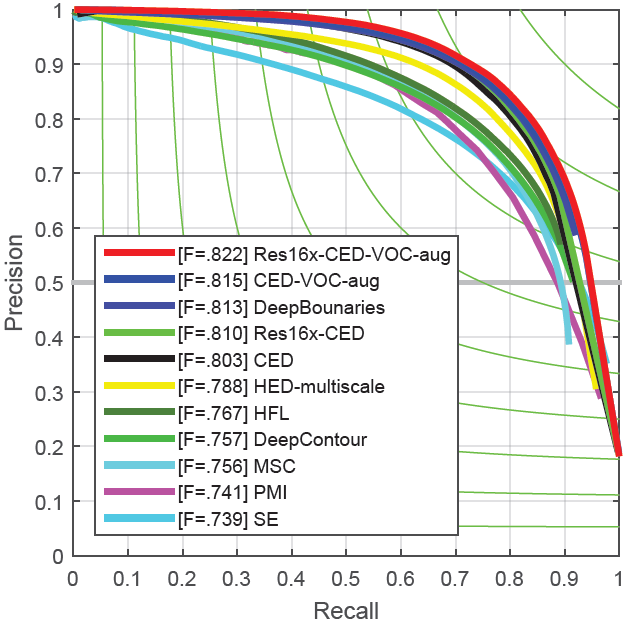}
\end{center}
\vspace{-1em}
   \caption{Precision/Recall curves of different methods on BSDS500 using standard evaluation criteria. CED is comparable to the previous best record by DeepBoundaries~\cite{kokkinos2015pushing}, which uses extra training data and post-processing steps. Simply augmenting training data as DeepBoundaries and without any post-processing, our CED-VOC-aug achieves state-of-the-art results. Specifically, using CED-VOC-aug and without complex post-processing steps, our Res16x-CED-VOC-aug surpasses all previous results. }
\label{fig:comp}
\end{figure}

\begin{figure*}[t]
\begin{center}
\includegraphics[width=0.88\linewidth]{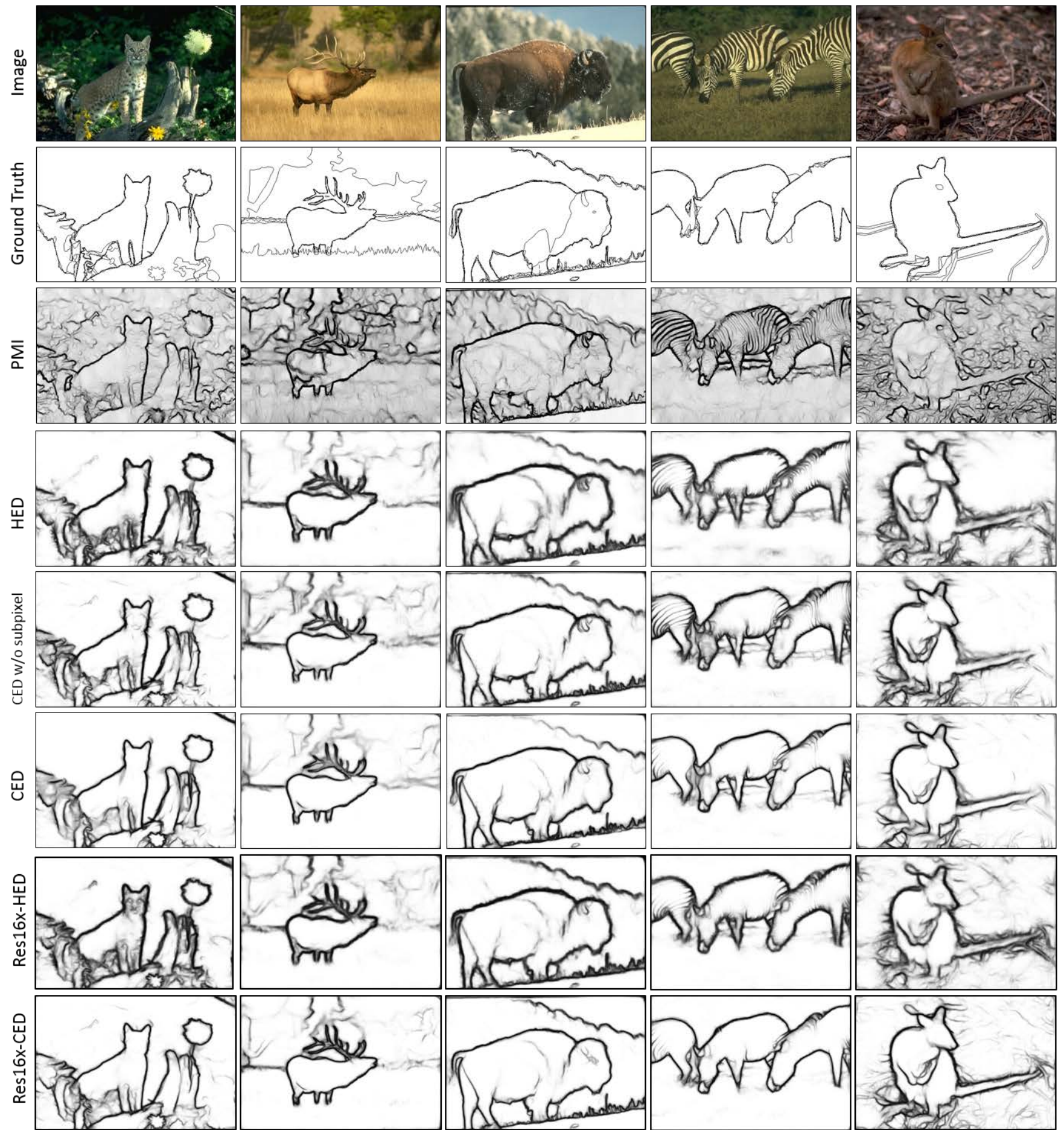}
\end{center}
   \caption{Visualization of edge detection results from different methods. First two rows show the original images and ground-truths edges. The next four rows include the raw edge maps (before NMS) of PMI, HED, CED-w/o-Subpixel, CED, Res16x-HED and Res16x-CED, respectively. Edge maps from CED is sharper than HED and cleaner than PMI. Similar results can be observed for both Res16x-HED and Res16x-CED.}
\label{fig:imgs}
\end{figure*}

\begin{table}
\begin{center}
\begin{tabular}{l|c|c|c}
Method&ODS & OIS&AP\\
\hline
Human&.8027&.8027&- \\
\hline
gPb-owt-ucm\cite{contour}&.726&.757&.696\\
SE-Var\cite{dollar2015fast} & .746 & .767 & .803 \\
PMI\cite{crisp}& .741 & .769 & .799\\
MES\cite{sironi2015projection}&.756&.776&.756\\

DeepEdge~\cite{kivinen2014visual}&.753&.769&.784\\
MSC~\cite{sironi2016multiscale}&.756&.776&.787\\
CSCNN~\cite{hwang2015pixel}&.756&.775&.798\\
DeepContour~\cite{shen2015deepcontour}&.757&.776&.790\\
HFL~\cite{bertasius2015high}&.767&.788&.795\\
HED~\cite{HED}& .788 & .808 & .840 \\
CEDN~\cite{ContourDetection2016}& .788 & .804 & - \\
RDS~\cite{liu2016learning}&.792&.810&.818\\
DeepBounaries~\cite{kokkinos2015pushing}&.813&.831&.866\\
\hline
CED & .803 & .820 & .871 \\
CED-VOC-aug & .815 & .833 & .889 \\
Res16x-CED & \textbf{.810} & \textbf{.829} & \textbf{.879} \\
Res16x-CED-VOC-aug & \textbf{.822} & \textbf{.840} & \textbf{.895} \\

\end{tabular}
\end{center}
\caption{Comparison to the state-of-art methods on BSDS500 dataset. We report ODS, OIS and AP scores. Without extra data and post-processing, our method is comparable to previous best (DeepBounaries). And with all bells and whistles (extra training data, post-processing, and better backbone network), our method outperform the best of the art.}
\label{tab:state}
\end{table}

\subsection{Results on BSDS500 Dataset}
We now present our results on BSDS500 dataset. These results are organized into two parts. First, we focus the ``correctness'' of the output edge map by using the standard evaluation criteria. Second, we evaluate the ``crispness'' of the edge maps by tightening the criteria.

\noindent \textbf{Correctness}. We compare our CED models to state-of-the-art methods on BSDS500 dataset using the standard evaluation criteria. Fig.~\ref{fig:comp} shows Precision-Recall curves of all methods for comparison. And Table~\ref{tab:state} summarizes their performance. Without multi-scale testing, CED-multi already achieves better results than the top-performing method~\cite{liu2016learning} in all $3$ metrics. Integrated with the multi-scale testing, CED achieves a further improvement, enhancing the ODS by $1.1\%$, OIS by $1.0\%$, and AP by $5.3\%$ in comparison to \cite{liu2016learning}. CED also significantly outperforms CEDN~\cite{ContourDetection2016}, which follows a form of encoder-decoder architecture~\cite{Noh2015Learning}. This result shows the superiority of the proposed top-down backward refinement architecture. This result also surpasses the human benchmark on the BSDS500 dataset with ODS $0.8027$. We note that the current record is from DeepBounaries~\cite{kokkinos2015pushing}, which used extra training samples (more than $10$K images in VOC) and post-processing steps of global grouping. After augmenting the standard BSDS500 training dataset with VOC images as~\cite{kokkinos2015pushing}, without any post-processing steps, this version of CED (CED-VOC-aug) gives better results with ODS $0.815$ (see Table~\ref{tab:state}).

Moreover, our improved version CED (Res16x-CED) is able to further boost the strong results of CED. Without using extra training data,  Res16x-CED boost the ODS to $0.810$, which is comparable to the best of the art (DeepBounaries~\cite{kokkinos2015pushing}). 
For our improved version of CED, Res16x-CED improves the performance to ODS $0.810$. After augmenting the standard BSDS500 training dataset with VOC images as~\cite{kokkinos2015pushing}, Res16x-CED gives an ODS of $0.822$, outperforming all other state-of-the-art results. The new form of CED is denoted as Res16x-CED-VOC-aug, as sown in Fig.~\ref{fig:comp} and Table~\ref{tab:state}.

\begin{figure*}[t]
\centering
	\subfigure[]{
    \includegraphics[width=0.9\linewidth]{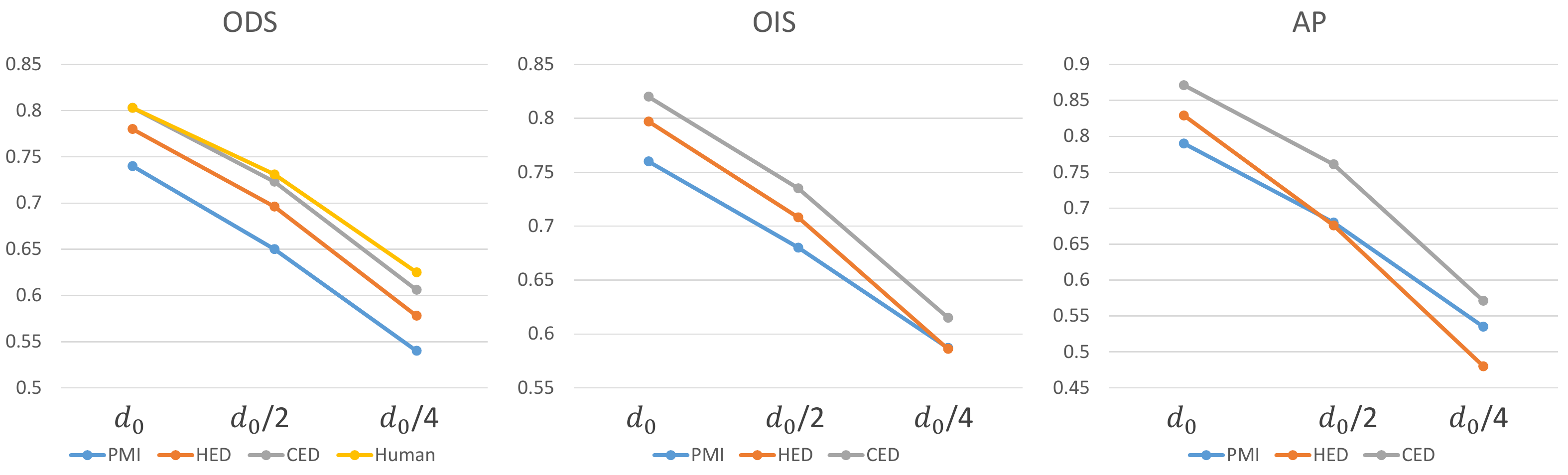}
	}
   	\subfigure[]{
    \includegraphics[width=0.9\linewidth]{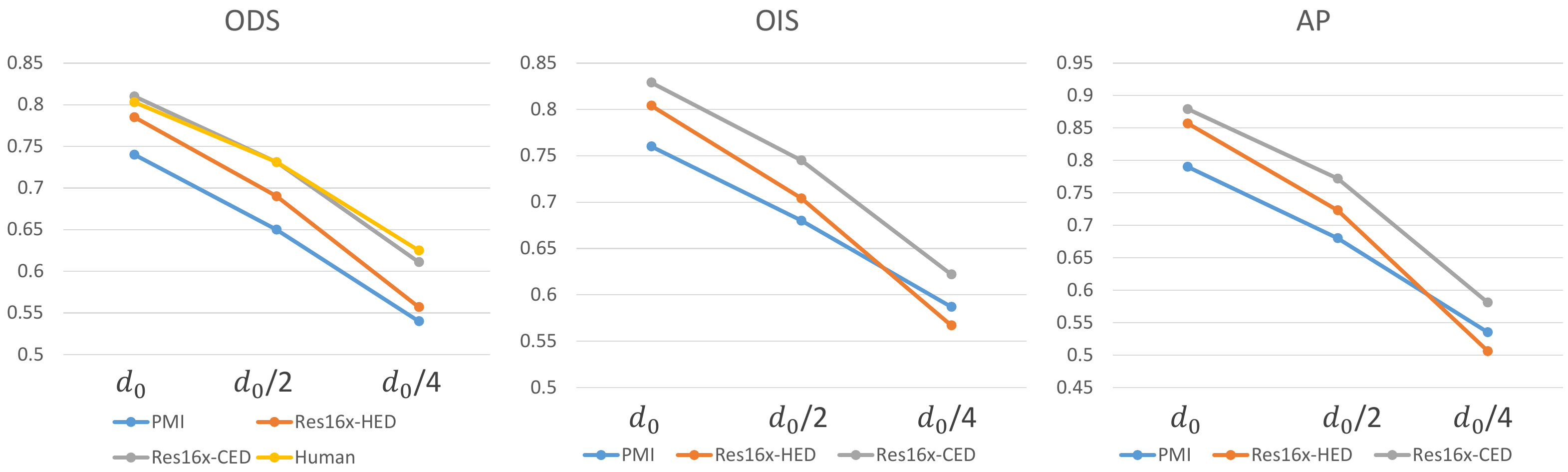}
	}
\vspace{-0.5em}
   \caption{We report the performance (ODS, OIS and AP) as a function of the maximal permissible distance $d$. (a) ``Crispness'' of boundaries generated by PMI, HED, CED and Human;  When $d$ decreases from $d_0$ to $d_0/4$, ODS gap between CED and HED increases from $2.3\%$ to $2.8\%$, the OIS gap increases from $2.3\%$ to $2.9\%$, and the AP gap increases from $4.2\%$ to $9.1\%$. Moreover, CED achieves ODS=$0.606$ at the setting of $d_0/4$, approaching human level performance ($0.625$), and outperforming the methods in~\cite{crisp,HED} by a large margin. (b) ``Crispness'' of boundaries generated by PMI, HED, Res16x-HED and Res16x-CED. When $d$ decreases from $d_0$ to $d_0/4$, the ODS gap between Res16x-CED and Res16x-HED increases from $2.5\%$ to $5.4\%$, the OIS gap increases from $2.5\%$ to $5.5\%$, and the AP gap increases from $2.2\%$ to $7.5\%$. Res16x-CED achieves an ODS of $0.611$ at $d_0/4$.}
\label{fig:localization}
\end{figure*}

\noindent \textbf{Crispness}. We further benchmark the ``crispness'' of edges from our methods. We report quantitative evaluation results by varying the matching distance $d$. This distance $d$ determines how flexible a output edge pixel can be matched to a nearby ground-truth edge. And thus decreasing $d$ will create a tightened evaluation criteria--the edge pixels has to be precisely localized. We evaluate CED and its improved version on the following settings of $d$: $d_0$, $d_0/2$, and $d_0/4$, where $d_0=0.0075$ is the standard criteria used in all previous benchmarks (4.3 pixels in a $321\times 481$ image). We compare the results of CED/Res16x-CED with HED/Res16x-HED~\cite{HED} (our baseline) and PMI~\cite{crisp} (designed for crisp edges). These results are plotted Fig.~\ref{fig:localization}(a-b).

The performance of all methods decreases as $d$ decreases. The gap between HED and PMI is getting closer with a smaller $d$. In contrast, the gap between CED/Res16x-CED and the baselines stays fairly consistent. In fact, the ODS gap between CED/Res16x-CED and HED/Res16x-HED increases from $2.3\%$/$2.5\%$ to $2.8\%$/$5.4\%$, the OIS gap increases from $2.3\%$/$2.5\%$ to $2.9\%$/$5.5\%$ and the AP gap increases from $4.2\%$/$2.2\%$ to $9.1\%$/$7.5\%$. This gap is more significant when using a strong backbone network (ResNet50). More importantly, CED achieves ODS = \textbf{0.606} at the setting of $d_0/4$, and Res16x-CED obtains ODS = \textbf{0.611}. This is a very challenging setting, as the edge pixel is only allowed to shift 1 pixel in the image plane. At this tight criteria $d_0/4$, our methods approach human level performance ($0.625$), and outperform the methods in~\cite{crisp,HED} by a large margin.

Finally, we visualize the edge maps from our methods and compare them against the baselines. Fig.~\ref{fig:imgs} shows a comparison of edge maps from PMI, HED and CED/Res16x-CED, before non-maximal suppression (NMS).  Even without the standard NMS, our method eliminates most blurry and noisy boundaries significantly. We observe that both CED and Res16x-CED are able to produce cleaner, thinner and crisper image boundaries. This is indeed the motivation of our work--designing a crisp boundary detector. While we did not claim to have solved this problem, we believe our work provides a promising step.

\begin{figure*}[t]
\begin{center}
\includegraphics[width=0.9\linewidth]{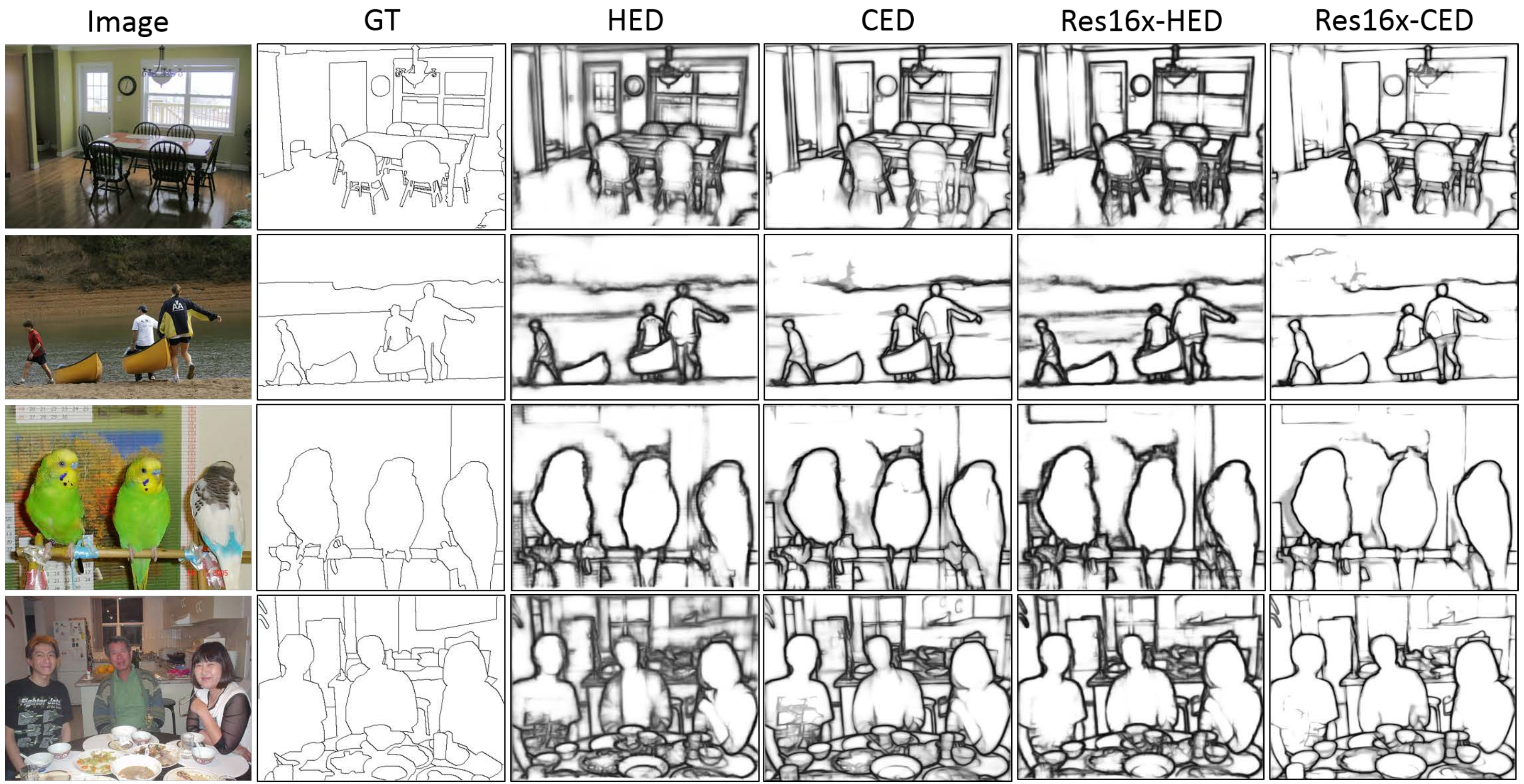}
\vspace{-0.5em}
\end{center}
   \caption{Visualization of edge detection results from different methods on PASCAL-Context dataset. First two columns show the original images and ground-truths edges. The next four rows include the raw edge maps (before NMS) of HED, CED, Res16x-HED, and Res16x-CED, respectively. Again, edge maps from CED models (CED/Res16x-CED) are sharper and cleaner than the ones from their HED versions.}
\label{fig:imgs_pascal}
\end{figure*}

\subsection{Boundary Detection on PASCAL-Context Dataset}
\label{subsec:pascal}
We then evaluate our method on PASCAL-Context dataset~\cite{mottaghi2014role}. As in~\cite{HED}, we set the matching tolerance as $d_{0} = 0.011$~(6.6 pixels in a $321\times481$ image) in evaluation, and report results on the its validation dataset. Our results are shown in Table~\ref{tab:boundary_pascal_context}. First, we present the results of models trained on on BSDS500~(HED-BSDS, CED-BSDS). This is an interesting test of cross-dataset generalization. In this case, HED-BSDS achieves an ODS of 0.570, while CED-BSDS boosts the performance to 0.639. Second, we report results of models (HED, CED, Res16x-HED, Res16X-CED) trained on PASCAL-Context training set. HED gives ODS 0.706, and CED improves the ODS by $2\%$. And both HED and CED when trained on PASCAL-Context are better than the same model trained on BSDS500 (around $10\%$). This result suggest a significant gap between these two dataset. Moreover, Res16x-HED improves over HED to ODS 0.722, and our Res16x-CED adds another 2.6\%. Our methods again achieve the best results. These results follow the same trend as our experiments on BSDS500 and thus provide further supports to our findings. Finally, we present the visualization of edge maps in Fig.~\ref{fig:imgs_pascal}. Again, Res16x-CED achieves sharp boundary maps and suppresses most background noise. These results not only confirm the effectiveness of our methods, but also demonstrate that our method can  scale to a much larger dataset (10x). 

Moreover, we also benchmark the ``crispness'' of CED and Res16x-CED on PASCSAL-Context dataset. Similar to our experiments on BSDS500, we vary the maximal permissible matching distance $d$. When the distance decreases from $d_0$ to $d_0/4$, the ODS gap between CED and HED increases from $2.0\%$ to $4.0\%$, the OIS gap increases from $2.5\%$ to $4.7\%$, and the AP gap increases from $4.6\%$ to $5.9\%$. The same trend holds for Res16x-CED and Res16x-HED. Their ODS gap increases from 2.6\% to 4.7\%, the OIS gap increases from 2.7\% to 5.1\%, and the AP gap increases from 3.5\% to 5.4\%. It can be also observed from Fig.~\ref{fig:imgs_pascal} that both CED and Res16x-CED achieve sharper and cleaner boundary maps than HED and Res16x-HED, respectively. These quantitative and qualitative results demonstrate that our methods can produce crisp edge maps on the more challenging PASCAL-Context dataset.

\begin{table}[t]
\begin{center}
\begin{tabular}{l|c|c|c}
Method&ODS & OIS&AP\\
\hline
HED-BSDS & .570 & .600 & .530\\
CED-BSDS & .639 & .673 & .640\\
\hline
HED & .706 & .725 & .732 \\
CED & \textbf{.726} & \textbf{.750} & \textbf{.778} \\
Res16x-HED & .722 & .741& .762 \\
Res16x-CED & \textbf{.748} & \textbf{.768} & \textbf{.797} \\
\end{tabular}
\end{center}
\caption{Boundary detection results on PASCAL-Context. HED achieves an ODS of 0.706, CED boosts the performance to 0.726. Res16x-HED gives 0.722 ODS, Res16x-CED further outperforms Res16x-HED by 2.6\%.}
\label{tab:boundary_pascal_context}
\end{table}

\section{Benefits of Crisp Boundaries}
\label{sec:benefits}
We further conduct experiments to demonstrate the benefits of crisp boundaries for other vision tasks. Specifically, We plug in the crisp edge from our methods into optical flow estimation, object proposal generation, and semantic segmentation. We evaluate the performance gain and discuss the results.

\subsection{Datasets and Benchmarks}
\label{sec:benefits_datasets}
We evaluate optical flow, object proposal and semantic segmentation using the following benchmarks. 

\noindent \textbf{Optical Flow}. Results are reported on MPI Sintel dataset~\cite{butler2012naturalistic}, a challenging optical flow evaluation benchmark obtained from animated sequences. We use the final version with photo-realistic rendering. As our method does not require the training of optical flow, we report Average Endpoint Error (AEE) on the training set as in~\cite{revaud2015epicflow,LiCVPR16edges}.

\noindent \textbf{Object Proposal}. We evaluate the results on Pascal VOC 2012 validation set~(VOC12 val set)~\cite{everingham2010pascal}. This set includes 1449 images from 20 common classes (e.g., train, person, sheep). Again we did not re-train the object proposal method and only replace the edge maps using our edge outputs (trained only on BSDS500). The Average Recall~(AR) with respect to the number of proposals is reported for the evaluation, as the standard metric in~\cite{ContourDetection2016,maninis2016convolutional}.

\noindent \textbf{Semantic Segmentation}. Results are evaluated on PASCAL-Context dataset~\cite{mottaghi2014role}. We evaluate on the most frequent 60 classes as~\cite{Chen2016DeepLab}. Pixel accuracy~(PA), mean pixel accuracy~(MPA), mean intersection over union~(Mean IOU) are reported as evaluation metrics.

\subsection{Optical Flow with Crisp Boundaries}
To further analyze the benefits of crisp edges, we make use of HED and CED results for optical flow estimation, a mid-level vision task. Optical flow is an important and challenging problem. The goal is to capture the motion information between neighboring frames in image sequence. We choose EpicFlow~\cite{revaud2015epicflow} as our optical flow estimation method. EpicFlow computes geodesic distance using an edge map, which is further used to interpolate sparse matches from~\cite{weinzaepfel:hal-00873592}. Thus, an accurate edge map is important for creating flow field that preserves object boundaries. 

\begin{figure}[t]
\centering
\includegraphics[width=0.95\linewidth]{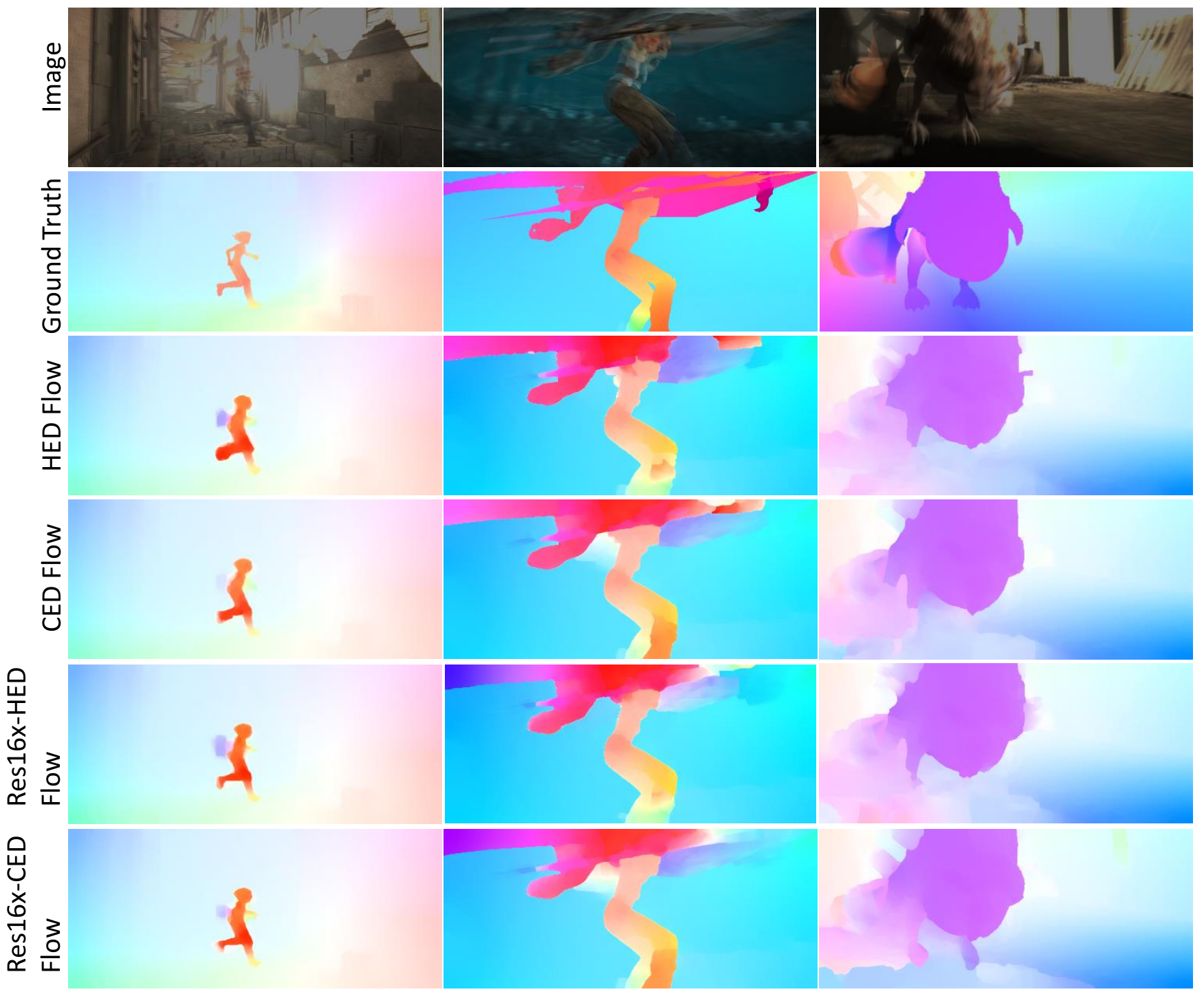}
   \caption{Visualization of optical flow estimation results with different edge maps. From top to bottom: mean of two consecutive images, ground-truth flow, and optical flow estimation results using edge maps of HED, CED, Res16x-HED, Res16x-CED. CED produces better motion details than HED, such as the leg of the girl in the second image. Similarly, Res16x-CED also gives more precise flow details than Res16x-HED.}
   \label{fig:flow}
\end{figure}

We compare CED and Res16x-CED results with the results of HED and Res16x-HED . In this case, we apply edge detectors trained with BSDS500 on Sintel dataset, and use the edge maps to interpolate the flow fields. The AEE on Sintel training set using CED is $3.570$ while HED gives $3.588$. Res16x-HED achieves an AEE of 3.573, Res16x-CED further reduces the AEE to 3.549. Fig.~\ref{fig:flow} shows the visualization of sample flow maps from Sintel. Again, CED achieves slightly more accurate flow results than HED. Similarly, Res16x-CED also captures more precise motion details than Res16x-HED. These results illustrate that optical flow estimation can benefit from a better localized edge map.

\begin{figure}[t]
\begin{center}
    \includegraphics[width=0.95\linewidth]{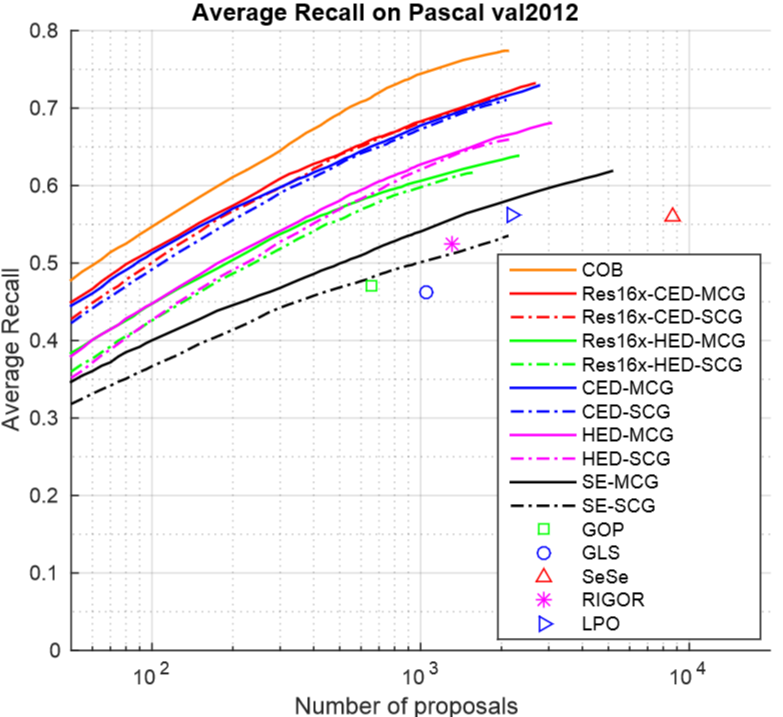}
\end{center}
\vspace{-0.5em}
   \caption{Results of object proposal generation on VOC12 val set. We evaluate three different edge detectors (SE, HED, CED) with two grouping methods (MCG, SCG), and report the Average Recall with respect to the number of proposals. CED-MCG outperforms HED-MCG by a large margin. Similarly, Res16x-CED-MCG surpasses Res16x-HED-MCG, giving 0.73 AR with $\sim$2650 proposals per image.}
\label{fig:proposals}
\end{figure}

\subsection{Object Proposal Generation with Crisp Boundaries}

We further demonstrate the benefit of the crisp edges for object proposal generation. Object proposal generation, another important mid-level vision task, is often the first step for other higher-level tasks, such as object detection. We choose the Multi-scale Combinatorial Grouping (MCG) and its single scale version (SCG) in~\cite{APBMM2014} to generate object proposals. With an input edge map, MCG builds a hierarchical grouping of contours to generate object proposals. The original MCG adopts the Structured Edge~(SE)~\cite{DollarICCV13edges} as the default edge detector. We simply replace the edge detector with HED~\cite{HED} and CED. We benchmark the combination of five edge detectors (SE, HED, CED, Res16x-HED, Res16x-CED) with both MCG and SCG. Note that these edge detectors are trained only on the BSDS500 dataset.

As in~\cite{ContourDetection2016,maninis2016convolutional}, we report the Average Recall~(AR) with respect to the number of proposals in Fig.~\ref{fig:proposals}. Both CED-MCG and HED-MCG achieve better results than SE-MCG. Specifically, HED-MCG achieves 0.68 AR with $\sim$3050 proposals per image, while CED-MCG boosts the performance to 0.73 AR with only $\sim$2750 proposals per image. Similarly, Res16x-HED-MCG obtains an AR of 0.64 with $\sim$2300 proposals per image, and Res16x-CED-MCG improves the AR to 0.73 with $\sim$2650 proposals per image. Similar results can be observed for the SCG with different edge detectors. Moreover, we also compare our methods with state-of-art methods, including COB~\cite{maninis2016convolutional}, GOP~\cite{Kr2014Geodesic}, GLS~\cite{Rantalankila2014Generating}, SeSe~\cite{Sande2011Segmentation}, RIGOR~\cite{Humayun2014RIGOR}, and LPO~\cite{Krahenbuhl2015Learning}. With edge detectors trained only on BSDS500 dataset, Res16x-CED-MCG narrows the performance gap between state-of-the-art COB~\cite{maninis2016convolutional}. These results demonstrate the benefit of crisp boundaries for high quality object proposals.

\begin{table}[t]
\begin{center}
\begin{tabular}{l|c|c|c}

Method &PA&MPA&Mean IOU \\
\hline
FCN-8s~\cite{Shelhamer2017Fully}& 67.0  & 50.7 & 37.8\\
CRF-RNN~\cite{Zheng2015Conditional}& n/a & n/a & 39.3\\
ParseNet~\cite{liu2015parsenet}& n/a &n/a & 40.4\\
PixelNet~\cite{bansal2017pixelnet}& n/a &n/a & 41.4\\
UoA-Context+CRF~\cite{Lin2016Efficient}&71.5&53.9&43.3\\
IFCN-8s~\cite{shuai2016improving}& 74.5  & 57.7 & 45.0\\
\hline
Deeplab& 70.7 & 54.5 & 42.6 \\
HED-BNF& 72.0 & 55.4 & 44.0 \\
CED-BNF& \textbf{72.1} & \textbf{55.5} & \textbf{44.2}\\
\hline
Res16x-HED-BNF& 72.1 & 55.5 & 44.1 \\
Res16x-CED-BNF& \textbf{72.5} & \textbf{55.7} & \textbf{44.5}\\


\end{tabular}
\end{center}
\caption{Semantic segmentation results on Pascal-Context validation set. With different edge detectors (HED, CED, Res16x-HED, Res16x-CED), we use BNF to further refine the initial results from DeepLab. For Mean IOU, HED-BNF improves the performance of Deeplab to 44.0, CED-BNF adds another 0.2. A similar trend holds when using ResNet backbone.}
\label{tab:BNF}
\end{table}

\begin{figure*}[t]
\begin{center}
    \includegraphics[width=0.9\linewidth]{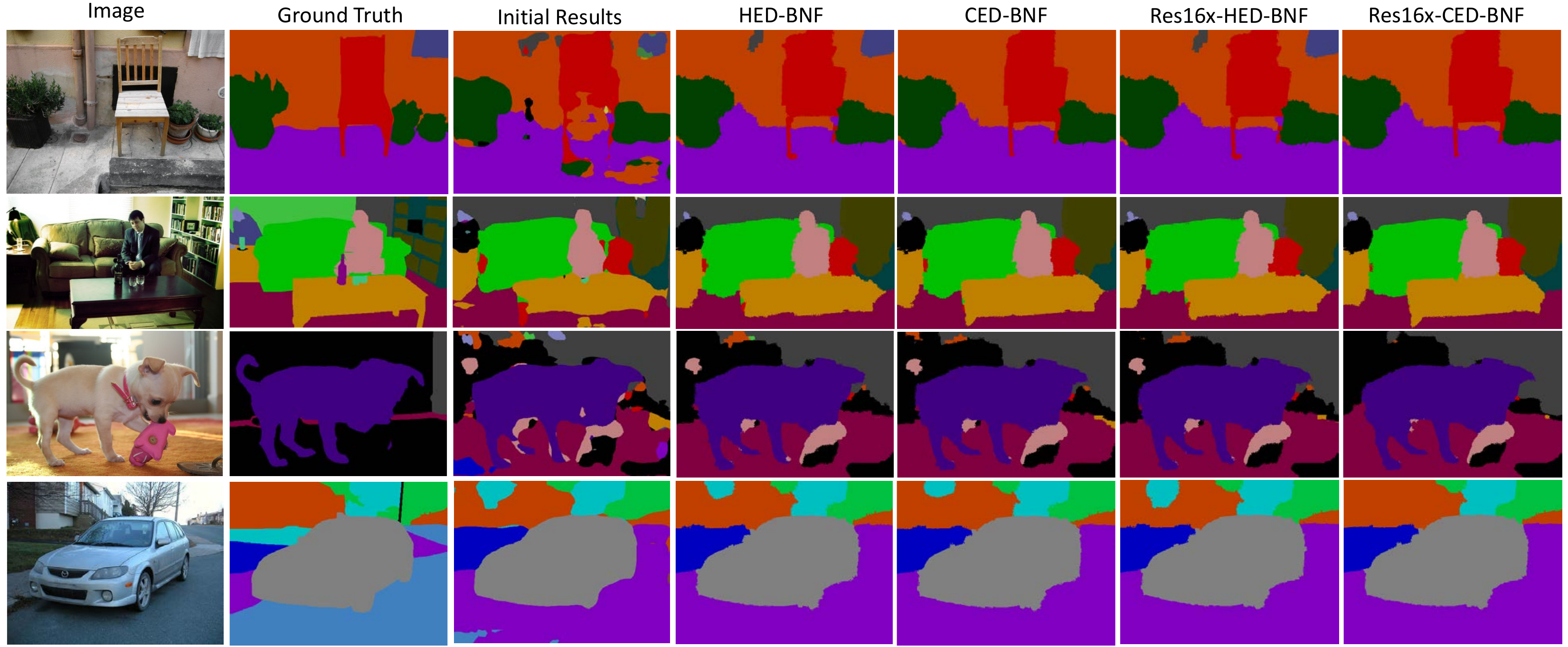}
\end{center}
   \caption{Sample semantic segmentation results on PASCAL-Context validation set. We use different edge detectors (HED, CED, Res16x-HED, Res16x-CED) for BNF that refines the initial results from DeepLab (third column). We also present the input image and the ground-truth maps. Our CED-BNF captures better details near object boundary than HED-BNF. Similar results can be observed between Res16x-CED-BNF and Res16x-HED-BNF.}
\label{fig:BNF}
\end{figure*}

\subsection{Semantic Segmentation with Crisp Boundaries}
Finally, we show that crisp boundaries can also help semantic segmentation. Semantic segmentation is an important high-level vision task. The task is to densely label each pixel with its semantic categories. Semantic segmentation has witnessed recent development with the help of fully convolutional networks~(FCNs)~\cite{Shelhamer2017Fully}. However, as discussed in~\cite{Bertasius2015Semantic}, due to the largely reduced resolution of successive down-sampling operations, segments generated by FCN-based methods are blob-like and did not capture the precise contour of the objects. 

To address this issue, Bertasius et al.\ \cite{Bertasius2015Semantic} proposed Boundary Neural Field~(BNF). BNF computes boundary-based pixel affinity functions, followed by a global optimization for segmentation. Specifically, BNF assigns a low similarity score between a pair of pixels that are separated by a strong boundary. And thus a high quality edge map will help the segmentation. We choose BNF for our experiment and replace the boundary map using our edge detectors' outputs. Our detectors (HED and CED) are trained using the PASCAL-Context training set, as in Section~\ref{subsec:pascal}. We use Deeplab~\cite{Chen2016DeepLab} to generate initial segmentation results, and experiment using BNF with different edge maps~(HED, CED, Res16x-HED, Res16x-CED) as the post-processing step. 

We report pixel accuracy~(PA), mean pixel accuracy~(MPA) and mean intersection over union~(Mean IOU) on Pascal-Context validation set in Table~\ref{tab:BNF}. The initial Deeplab gives a Mean IOU of 42.6, the post-processed results by BNF with HED and CED edge maps both improve the performance. HED-BNF achieves a Mean IOU of 44.0, CED-BNF improves the performance to 44.2. Similarly, Res16x-HED-BNF gets 44.1 Mean IOU, Res16x-CED-BNF further boosts the performance to 44.5, which is comparable to state-of-arts~\cite{shuai2016improving}. Fig.~\ref{fig:BNF} shows the initial segmentation results of Deeplab, post-processed results with HED-BNF, CED-BNF, Res16x-HED-BNF, and Res16x-CED-BNF. All the post-processed results enhance the initial segmentation results. Particularly, with crisp boundaries, CED-BNF preserves more details around object boundary than HED-BNF. Similarly, compared to Res16x-HED-BNF, Res16x-CED-BNF also delineates more precise segmentation results. These results show the benefit of crisp boundaries for semantic segmentation. Finally, Fig~\ref{fig:BNF} shows sample results of the baseline and our methods. CED-BNF is able to improve the initial segmentation results by better capturing details around the object contour.

\section{Conclusion}
In this paper, we demonstrated that ConvNet based edge detector tends to generate edge maps which are not well aligned with image boundaries. We discussed the reason behind the issue and proposed a novel architecture that largely improved the localization ability of ConvNet based edge detectors. Our detector achieved promising performance on BSDS500 and PASCAL-Context, outperforming the state-of-the-art methods when using more strict maximum tolerance setting. More importantly, we verified the benefits of crisp edge map for optical flow estimation, object proposals generation and semantic segmentation, covering a range of mid-level and high-level vision tasks. To summarize, our work revisited the classical problem of edge detection in computer vision. We hope that our work will provide a reflection of the recent victory of ConvNet in vision tasks. While standard quantitative results seem to be improving over time, the fundamental vision problems remain challenging---it is probably the right time to revisit our problem definition and evaluation criteria.

\section*{Acknowledgement}
This project was partial supported by the National Key Research and Development Program of China (Grant No.\ 2016YFB1001005), the National Natural Science Foundation of China (Grant No.\ 61673375 and Grant No.61602485), and the Projects of  Chinese Academy of Science (Grant No.\ QYZDB-SSW-JSC006 and Grant No.\ 173211KYSB20160008).

{\small 
\bibliographystyle{IEEEtran}
\bibliography{egbib}
}

%

\begin{IEEEbiography}[{\includegraphics[width=1.25in,height=1.25in,clip,keepaspectratio]{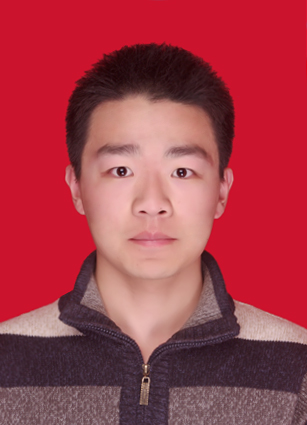}}]{Yupei Wang} received his BSc from the Henan University in 2014.
    He is currently working toward the PhD degree in the National Laboratory of Pattern Recognition (NLPR), Institute of Automation, Chinese Academy of Sciences (CASIA), Beijing, China. He is also with the University of Chinese Academy of Sciences.
    His research interests include computer vision, deep learning and semantic segmentation.
\end{IEEEbiography}

\begin{IEEEbiography}[{\includegraphics[width=1.25in,height=1.25in,clip,keepaspectratio]{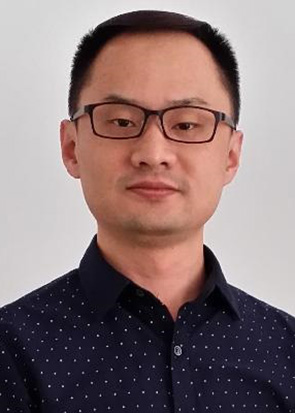}}]{Xin Zhao} received the B.E. degree from Anhui University of Technology (AHUT) in 2006 and Ph.D. degree from University of Science and Technology of China (USTC) in 2013. In July 2013, he joined the Institute of Automation, Chinese Academy of Sciences (CASIA), where he is currently an Associate Professor. Dr. Zhao has published research papers in the areas of computer vision and pattern recognition at international journals and conferences such as CVIU, CVPR, ICCV, IJCAI, ICPR, ICIP and ACCV. In 2011, he won the IAPR (the International Association of Pattern Recognition) Best Student Paper Award. His current research interests include pattern recognition, computer vision and machine learning.
\end{IEEEbiography}

\begin{IEEEbiography}[{\includegraphics[width=1.25in,height=1.25in,clip,keepaspectratio]{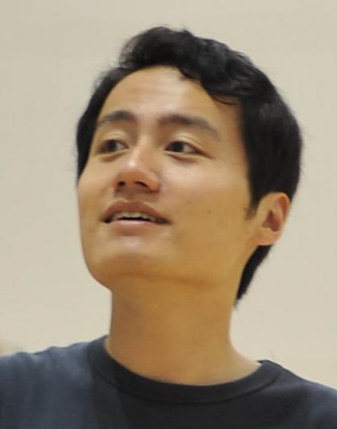}}]{Yin Li} is an assistant professor at the University of Wisconsin--Madison. Previously, he obtained his PhD from the Georgia Institute of Technology and was a Postdoctoral Fellow at the Carnegie Mellon University. Dr.\ Li's research interests lie at the intersection of computer vision and mobile health. Specifically, he creates methods and systems to automatically analyze first person videos, known as First Person Vision (FPV). He has particular interests in recognizing the person's activities and developing FPV for health care applications. He is the co-recipient of the best student paper awards at MobiHealth 2014 and IEEE Face and Gesture 2015. His work had been covered by MIT Tech Review, WIRED UK and New Scientist.
\end{IEEEbiography}




\begin{IEEEbiography}[{\includegraphics[width=1.25in,height=1.25in,clip,keepaspectratio]{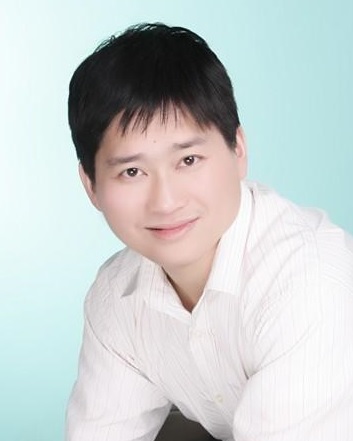}}]{Kaiqi Huang} received his BSc and MSc from Nanjing University of Science Technology, China and obtained his PhD degree from Southeast University. He has worked in National Lab. of Pattern Recognition (NLPR), Institute of Automation, Chinese Academy of Science, China and now he is a professor in NLPR. His current research interests include visual surveillance, digital image processing, pattern recognition and biological based vision and so on.
\end{IEEEbiography}




\end{document}